\documentclass[runningheads]{llncs}
\usepackage[T1]{fontenc}
\usepackage{graphicx}
\usepackage[hmarginratio=1:1, hmargin=1.75in]{geometry}
\usepackage[cleardoublepage=plain]{scrextend}
\usepackage{nicematrix}
\usepackage[utf8]{inputenc}
\usepackage{array}
\usepackage{csquotes}
\usepackage{soul}
\usepackage{float} 
\usepackage[hidelinks]{hyperref}
\usepackage{comment}
\usepackage{todonotes}
\usepackage{fancyhdr}
\usepackage{lipsum}  
\usepackage{hyperref}
\usepackage{amsmath,amssymb,bm}
\usepackage{cancel}
\usepackage{amsfonts}
\usepackage{makecell}
\usepackage{ctable}
\usepackage[ruled,noend]{algorithm2e}
\usepackage{todonotes}

\begin{document}
\title{Job Shop Scheduling via Deep Reinforcement Learning: a Sequence to Sequence approach}

%
\author{Giovanni Bonetta\thanks{Equal contribution}\orcidID{0000-0003-4498-1026} \and
Davide Zago\inst{\star}\orcidID{0000-0003-1112-3543} \and \\
Rossella Cancelliere\orcidID{0000-0002-9120-3799} \and 
Andrea Grosso\orcidID{0000-0002-9926-2443}}
\authorrunning{Bonetta et al.}
%
\institute{Department of Computer Science, University of Turin, 10149 Turin 
\email{\{giovanni.bonetta,rossella.cancelliere,andrea.grosso\}@unito.it} 
\email{zago@di.unito.it}}
%
\maketitle              
\begin{abstract}
Job scheduling is a well-known Combinatorial Optimization problem with endless applications. Well planned schedules bring many benefits in the context of automated systems: among others, they limit production costs and waste. Nevertheless, the NP-hardness of this problem makes it essential to use heuristics whose design is difficult, requires specialized knowledge and often produces methods tailored to the specific task.
This paper presents an original end-to-end Deep Reinforcement Learning approach to scheduling that automatically learns dispatching rules. Our technique is inspired by natural language encoder-decoder models for sequence processing and has never been used, to the best of our knowledge, for scheduling purposes.
We applied and tested our method in particular to some benchmark instances of Job Shop Problem, but this technique is general enough to be potentially used to tackle other different optimal job scheduling tasks with minimal intervention.
Results demonstrate that we outperform many classical  approaches exploiting priority dispatching rules and show competitive results on state-of-the-art Deep Reinforcement Learning ones.

\end{abstract}
\section{Introduction}
\label{sec:intro}

{\let\thefootnote\relax\footnotetext{\\Published in: Proceedings of the 17\textsuperscript{th} Learning and Intelligent Optimization Conference (LION17), Nice 4-8 June 2023. Complete version is published by Springer LNCS edt.}
}

\textit{Job Shop Problem} (JSP) is a well-known Combinatorial Optimization problem fundamental in various automated systems applications such as manufacturing, logistics, vehicle routing, telecommunication industry, etc... In short, some jobs with predefined processing constraints have to be assigned to a set of heterogeneous machines, to achieve the desired objective (e.g. minimizing the flowtime).
Due to its NP-hardness, finding exact solutions to the JSP is often impractical (or impossible, in many real-world scenarios), but many tasks can be effectively addressed through heuristics \cite{glover1998CO,Haupt1989ASO} or approximate methods \cite{10.1007/10719839_7}, that represent the most suitable choice for large-scale problems, providing near optimal solutions with acceptable computational times.

Heuristic algorithms are classified as constructive or as local search methods. Constructive heuristics assemble the solution with an incremental process: at each step, the choice of the next element in the solution is made by examining some local information of the problem, and once one variable has been fixed it's not reconsidered. 
\textit{Priority Dispatching Rules} (PDRs) \cite{Haupt1989ASO} belong to the category of constructive approximate methods: each operation is allocated in a dispatching sequence following a monotonic utility measure.

The use of dispatching rules emerged very early in the scheduling area, and it is well established by now. Most dispatching rules are known to be less than a match for modern, sophisticated heuristic optimization techniques (e.g. simulated annealing, tabu search, etc); despite this, they are still commonly used in many practical contexts because they are 
considered quick, flexible and adaptable to many situations. Besides, PDRs are widely used in real-world scheduling systems because they are intuitive and easy to implement. As a result, optimization literature is rich of PDR methods for the JSP \cite{Sels2012PDR}, even if it is well known that designing an effective PDR is time-consuming and requires a substantial domain knowledge. 

A possible solution is the automation of the process of designing dispatching rules: recent works on learning algorithms for Combinatorial Optimization (see \cite{BENGIO2021405} for a survey) show that Deep Reinforcement Learning (RL) could be an ideal technique for this purpose, and in particular that it can be considered a potential breakthrough in the construction of heuristic methods for the JSP \cite{cunha2018deep}.
Reinforcement Learning \cite{sutton2018reinforcement} is a subfield of Machine Learning (ML) that experienced a great development in recent years, mainly thanks to the contribution of Deep Learning.

The main idea of this paper is to treat the JSP as a sequence to sequence process: inspired by deep learning natural language models we propose a Deep Reinforcement Learning approach that, exploiting the encoder-decoder architecture typical of language, automatically learns robust dispatching rules. This leads us to consider PDRs as a reasonable match for deep RL-based optimization techniques that, it should be remembered, despite of the huge amount of works appearing on the subject, are still in their infancy.

Our method is able to learn dispatching rules with higher performance than traditional ones, e.g. \textit{Shortest Processing Time} (SPT), \textit{Most Work Remaining} (MWKR). On top of that, our approach shows competitive results against state-of-the-art Deep RL methods when tested on small and medium sized JSP benchmark instances. Besides, it shows a high degree of flexibility: \textit{Flow Shop Problem} (FSP) instances can also be solved, and minimal modifications to the model would allow solving \textit{Open Shop Problem} (OSP). 

Since the model requires sequences as inputs and outputs we design an appropriate, yet compact and easily interpretable encoding for JSP instances and solutions. Besides, thanks to a tailored masking procedure, the model outputs a permutation of job operations (virtually a priority list) that respects precedence constraints and can be mapped to a \textit{schedule}, i.e. the association of each operation to a specific starting time.

The rest of this paper is organized as follows: \autoref{sec:sota} contains an overview of related works concerning neural and Deep Reinforcement Learning methods for Combinatorial Optimization (CO). \hyperlink{sec:foundations}{Section 3} provides the definition of Markov Decision Process (MDP) and the theoretical foundations of our model. \hyperlink{sec:JSP}{Section 4} introduces the mathematical notation which formalizes the JSP. \hyperlink{sec:Method}{Section 5} describes our technique for sequence encoding, the neural architecture used and the proposed masking mechanism, the experimentation details and the results obtained.

\section{Related works}
\label{sec:sota}

Before Deep RL gained the popularity it has today, many ML-based approaches have been applied to CO (see \cite{smith1999neural} for an in-depth overview), such as assignment problems, cutting stock and bin packing problems, knapsack problems, graph problems, shortest path problems, scheduling problems, vehicle routing problems and the Travelling Salesman Problem (TSP).

\vspace{20px}

\textit{...OMISSIS...}

\section{Mathematical foundations}
\label{sec:foundations}

RL substantially differs from other ML paradigms since it's concerned with how an agent learns to act in an environment: agents' behavior is optimized through a training phase, requiring the definition of a Markov Decision Process \cite{bellman1957markovian}, focused on the maximization of a cumulative expected reward collected through a sequence of actions. 

An MDP is a mathematical framework used to formalize a general decision making process involving a single agent acting in an environment. It is defined by a tuple $M=(S,A,R,T,\gamma,H)$ where:

\begin{itemize}
    \item $S$ - \textit{state space}. 
    
    It is the set of all the possible representations $s$ of the environment and of the agent's internal state at a given time.
    \item $A$ - \textit{action space}. 
    
    It is the set of all the possible actions $a$ the agent can perform.
    \item $R$ - \textit{reward function} $R : S \times A \times S \rightarrow \mathbb{R}$. 
    
    It is the reward given to the agent after doing action $a$ in state $s$ and landing in state $s'$.
    \item $T$ - \textit{transition function} $T(s'|s,a)$. 
    
    It is the transition probability from state $s$ to $s'$ given that action $a$ has been performed.
    \item $\gamma$ - \textit{discount factor}. 
    
    It weights the rewards of future actions. $\gamma \in [0,1]$.
    \item $H$ - \textit{time horizon}. 
    
    It is the maximum number of transition that can occur before the decision process is halted.
\end{itemize}

\noindent The objective of RL is to maximize the expected return of the sequence of actions performed by the agent. Each action is sampled from a \textit{stochastic policy} $\pi(a|s)$, with $a \in A$ and $s \in S$, i.e. a probability distribution over the set of actions given a particular state.

\subsection{Policy gradient algorithms}
\label{sec:policy_gradient}

\textit{Policy gradient} (or \textit{policy optimization})  methods \cite{sutton2018reinforcement} are widely used in Deep RL research and directly optimize the stochastic policy $\pi_\theta$, which is approximated by a neural network with parameters $\theta$.

By taking actions in the environment, the agent defines trajectories. A \textit{trajectory} $\tau$ (alternatively \textit{episode} or \textit{rollout}) is a sequence of states and actions $(s_0,a_0,s_1,a_1,...,s_{H-1},a_{H-1},s_H)$ and it has a return $R(\tau)$ associated to it:

\begin{equation}
R(\tau) = \sum_{t=0}^H R(s_t,a_t,s_{t+1})
\end{equation}

\noindent $R(\tau)$ is called \textit{finite-horizon undiscounted return} since it's defined with horizon $H$. Moreover, the probability of a trajectory given the policy is:

\begin{equation}
P_{\theta}(\tau) = \rho(s_0) \prod_{t=0}^H T(s_{t+1}|s_t, a_t) \pi_\theta(a_t|s_t)
\end{equation}

\noindent where $\rho(s_0)$ is the a priori probability of state $s_0$.

Given the parameterized stochastic policy $\pi_\theta$, the learning objective is the maximization of the expected return w.r.t. a set of trajectories:

\begin{equation}
\max_\theta J(\pi_\theta), \ \text{where} \ J(\pi_\theta) = \mathop{\mathbb{E}}_{\tau \sim \pi_\theta} [R(\tau)]
\end{equation}

\noindent Considering a policy optimized with gradient ascent,
the quantity $\nabla_\theta J(\pi_\theta)$ is called \textit{policy gradient} and the following equation holds:

\begin{equation}\label{eq:policy_gradient}
\nabla_\theta J(\pi_\theta) = \mathop{\mathbb{E}}_{\tau \sim \pi_\theta} \left[ \sum_{t=0}^H \nabla_\theta \log \pi_\theta(a_t|s_t) R(\tau) \right]
\end{equation}

\noindent This leads to the \textit{REINFORCE} algorithm (\autoref{alg:reinforce}), also known as \textit{Vanilla policy gradient}, for optimizing policies, first proposed by Williams in \cite{williams1992simple}.

\vspace{10px}

\begin{algorithm}[H]
\SetAlgoNoLine
\DontPrintSemicolon
\KwIn{MDP $M=(S,A,T,R,\gamma,H)$}
\KwOut{policy $\pi_{\theta_k}$}
    $\theta_0 \gets$ \textsc{initial-parameters()}\;
    \For{$k\in(0,1,2,...)$}{
        $\mathcal{D} \gets $ \textsc{collect-trajectories()}\;
        $g_k \gets \frac{1}{|\mathcal{D}|} \sum_{\tau \in \mathcal{D}} \sum_{t=0}^H \nabla_\theta \log \pi_\theta(a_t|s_t) R(\tau)$ \hfill *\{policy gradient\}
        \hspace*{-8px} $\theta_{k+1} \gets \theta_k + \alpha g_k$ \hfill**\{gradient ascent step\}
    }%
\Return{$\pi_{\theta_k}$}
\caption{REINFORCE}%
\label{alg:reinforce}%
\end{algorithm}

\vspace{10px}

\noindent Equation $*$ is the estimation of the policy gradient over the set of trajectories $\mathcal{D}$. Statement $**$ -- i.e. the gradient ascent update rule --- can be substituted with the update rule of a different optimization algorithm, e.g. Adam.

Unfortunately the \textit{unbiased policy gradient} $g_k$ suffers from high variance which hinders performance and learning stability. This can be addressed through the use of baselines, terms that only depend on the current state and are subtracted from the reward. \autoref{eq:baselines} is the policy gradient updated with a generic baseline term.

\begin{equation}\label{eq:baselines}
\nabla_\theta J(\pi_\theta) = \mathop{\mathbb{E}}_{\tau \sim \pi_\theta} \left[ \sum_{t=0}^H \nabla_\theta \log \pi_\theta(a_t|s_t) \left( \sum_{t=0}^H R(s_t,a_t,s_{t+1}) - b(s_t)\right) \right]
\end{equation}

\section{The Job Shop optimization problem: notation}
\label{sec:JSP}

Scheduling is a decision-making process consisting in the allocation of resources to tasks over a given time period, with the additional constraint of optimizing one (or more) objective functions. The JSP is one of the most studied scheduling problems, along with the Open Shop and the Flow Shop Problems. A $n \times m$ JSP instance is characterized by:
\begin{itemize}
\item $n$ \textit{jobs}  $J_i,$ with $i\in\{0,...,n-1\}$, each one consisting of $m$ operations (or tasks) $O_{ij}$, with $j \in \{0,...,m-1\}$.

\item $m$ \textit{machines} $M_{ij} $, with $j\in\{0,...,m-1\}$. $M_{ij}$ identifies the machine required to execute the $j$-th operation of job $i $.
\end{itemize}

\noindent We denote the execution time of an operation $O_{ij}$ with $p_{ij}$; an operation execution cannot be interrupted and each operation of a given job must be executed on a different machine. A JSP solution is represented by a schedule.

As an example, let us consider the JSP instance represented in \autoref{tab:jsp_instance}. In this case there are three jobs $J_i$, with $i\in\{0,1,2\}$, and four operations $O_{ij}$ for the $i$-th job, with $j \in \{0,1,2,3\}$. Operation $O_{ij}$ must be executed on machine $M_{ij} \in \{0,1,2,3\}$ and has processing time $p_{ij}$. 

\vspace*{-10px}

\begin{table}[h]
    \centering
    \begin{tabular}{|c|cccc|}
        \hline
        $M_{ij},p_{ij}$ & $O_{*0}$ & $O_{*1}$ & $O_{*2}$ & $O_{*3}$ \\
        \hline
         $J_0$ & (0, 4) & (2, 2) & (1, 6) & (3, 2) \\
        $J_1$ & (0, 4) & (3, 5) & (2, 7) & (1, 8) \\
        $J_2$ & (2, 6) & (0, 4) & (1, 3) & (3, 1) \\
         \hline
    \end{tabular}
    \caption{Example of a $3 \times 4$ JSP instance.}
    \label{tab:jsp_instance}
\end{table}

\vspace*{-20px}

\begin{figure}
    \centering
    \includegraphics[width=340px]{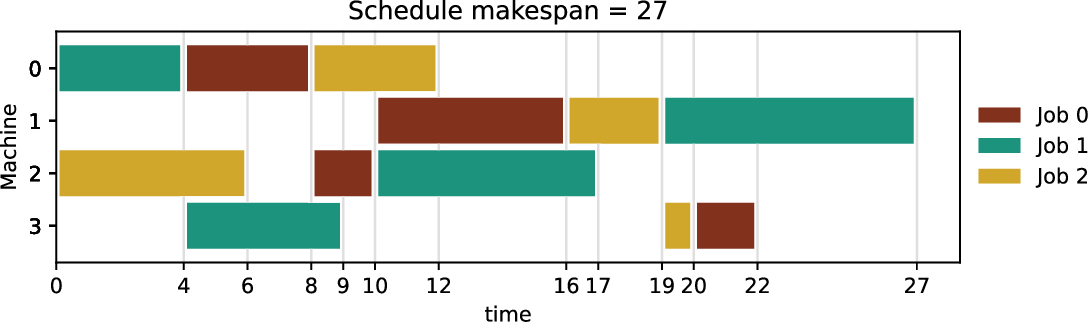}
    \caption{One possible schedule for the JSP instance in \autoref{tab:jsp_instance}.}
    \label{fig:sch2}
\end{figure}

\noindent A useful tool for visualizing a schedule is Gantt charts \cite{ganttgraphical}. 
 \autoref{fig:sch2} represents the Gantt chart for a possible schedule of the JSP instance represented in \autoref{tab:jsp_instance}. 

\noindent The optimal solution of a JSP is the schedule that minimizes the makespan $C_{max}$, where $C_{max} =\smash{\displaystyle\max_{i}}\:\:C_i$, and $C_i$ is the completion time of the $i$-th job.

\section{Our Sequence to Sequence approach to the JSP}
\label{sec:Method}

\begin{figure}[H]
    \centering
    \includegraphics[width=320px]{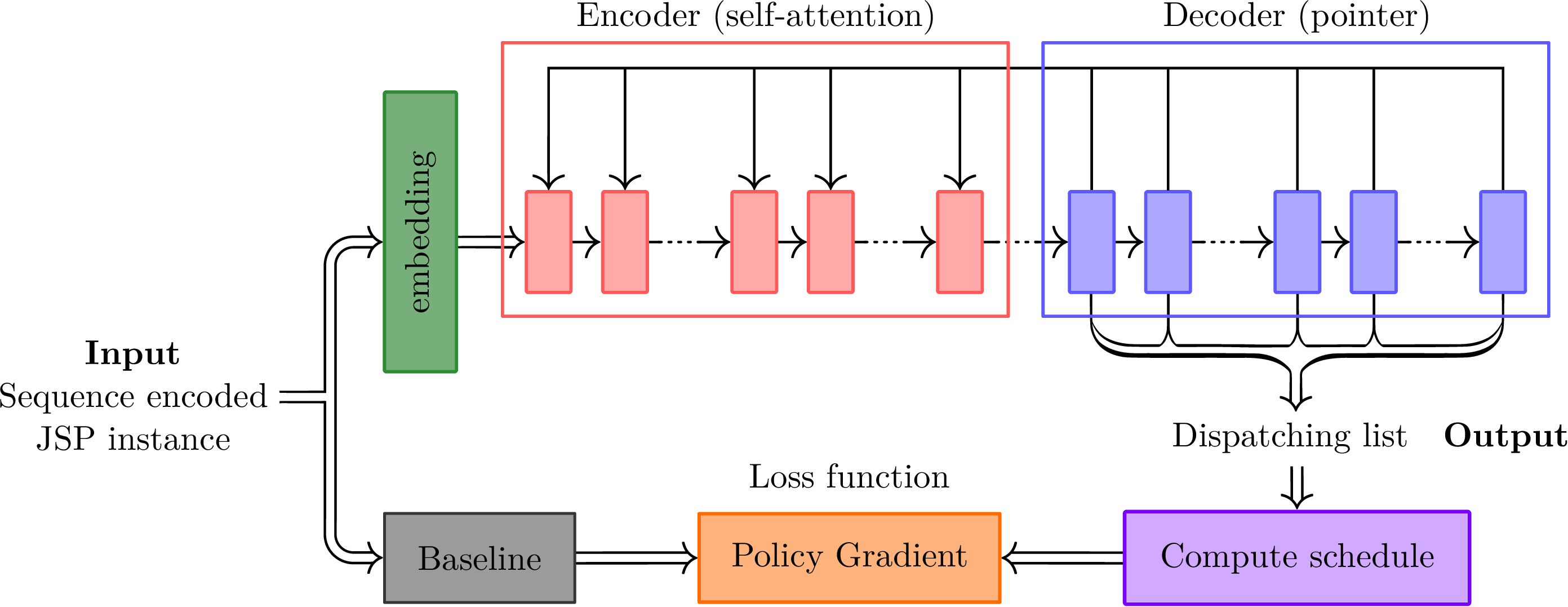}
    \caption{Our encoder-decoder architecture for scheduling problems.}
    \label{fig:model_architecture}
\end{figure}

The main novelty we present is a sequence-based Deep RL approach applied to the JSP. Inspired by \cite{bello2017neural} and \cite{kool2019attention} we make use of a deep neural network used for NLG applications and we train it in a RL setting. Such model (see \autoref{fig:model_architecture}) combines a self-attention based encoder and a Pointer-Network decoder \cite{vinyals2015pointer}. In order to apply it to the JSP, we formulate a sequence-based  encoding of input and output, and design an appropriate masking mechanism to generate feasible solutions. Our code is available on Github\footnote{Github repository: \url{https://github.com/dawoz/JSP-DeepRL-Seq2Seq}}.

\vspace*{-5px}

\vspace*{-25px}

\subsection{Sequence encoding}
The input (i.e. problem instance) and the model's output (i.e. solution) need to be encoded as sequences in order for the model to process them correctly.

We consider both the input and the output as sequences of operations and we define a 4-dimensional feature vector $\mathbf{o}_k$ for each operation $O_{ij}$ as follows:

\begin{equation}
\mathbf{o}_k = \left[
i \ \ j \ M_{ij} \ p_{ij}
\right]
\qquad \text{with} \ k = m \cdot i  + j
\end{equation}

\noindent where $i$ is the index of the $i$-th job and $j$ the index of its $j$-th operation. Consider a JSP instance $S$ with $n$ jobs $J_i$ ($i \in \{0,...,n-1\}$) and $m$ operations $O_{ij}$ for job $J_i$ ($j \in \{0,...,m-1\}$) with required machine $M_{ij} \in \{0,...,m-1\}$ and execution time $p_{ij}$. $S$ can be expressed with the following sequence encoding $S^{\text{seq}}$:

\begin{equation}
\resizebox{.67\hsize}{!}{
$
S^{\text{seq}} = 
\left[\begin{array}{ll}
\mathbf{o}_0\\
\mathbf{o}_1\\
\vdots  \\
\mathbf{o}_{m-1}\\
\mathbf{o}_m\\
\vdots  \\
\mathbf{o}_{(m-1)(n-1)}\\
\end{array}\right]
= \overbrace{\left[\begin{array}{llll}
0 & 0 & M_{00} & p_{00}\\
0 & 1 & M_{01} & p_{01}\\
\vdots & \vdots & \vdots & \vdots \\
0 & m-1 & M_{0m-1} & p_{0m-1}\\
1 & 0 & M_{10} & p_{10}\\
\vdots & \vdots & \vdots & \vdots \\
n-1 & m-1 & M_{n-1m-1} & p_{n-1m-1}\\
\end{array}\right]}^{i \qquad \ \; j \qquad \ \ M_{ij} \quad \qquad p_{ij} \qquad \ \ }
$
}
\end{equation}

\noindent The matrix $S^{\text{seq}}$ just defined determines which jobs/operations have to be handled: establishing an order in which to execute them allows to identify a schedule.

Besides, a correct encoding of the model's output sequence implies that if for job $J_i$ operation $O_{ij}$ must be executed before $O_{ik}$, then the vectors of the operations in the output sequence must occur in the same order (not consecutive, in general).

\noindent In order to comply with these requests, the sequence encoding for the output $L^{\text{seq}}$ of the model has the following form:

\begin{equation}
L^{\text{seq}} = P S^\text{seq}=\left[\mathbf{o}'_0 \hdots  \mathbf{o}'_{(n-1)(m-1)}\right]^T
\end{equation}

\noindent where $P$ is a permutation matrix suitable for obtaining a matrix $L^\text{seq}$ which encodes a feasible JSP solution. 

The condition under which this occurs is explained in the following definition:

\begin{definition}[Feasible sequence encoded JSP Solution]
\label{def:validJSP}

\noindent Let $O_{ij}$ and $O_{ik}$ be the operations of the $i$-th job with $j < k$,\\and $\mathbf{o}'_{s} = \left[ i \ \ j \  M_{ij} \ p_{ij} \right]$, $\mathbf{o}'_{r} = \left[ i \ \ k \  M_{ik} \ p_{ik} \right]$.


\noindent The matrix $L^{\text{seq}}$ is the sequence encoding of a feasible schedule iff the permutation $ P $ is such that $ s < r$, for all $s$ and $r$ in $\{0,...,(n-1)(m-1)\}$ (i.e. the order of operations for job $i$ defined in $S^{\text{seq}}$ is preserved).
\end{definition}

\noindent As an example the sequence encodings of the JSP instance in \autoref{tab:jsp_instance} are the following:

$$
S^{\text{seq}} = \overbrace{\begin{bmatrix}
0 & 0 & 0 & 4\\
0 & 1 & 2 & 2\\
0 & 2 & 1 & 6\\
0 & 3 & 3 & 2\\
1 & 0 & 0 & 4\\
1 & 1 & 3 & 5\\
1 & 2 & 2 & 7\\
1 & 3 & 1 & 8\\
2 & 0 & 2 & 6\\
2 & 1 & 0 & 4\\
2 & 2 & 1 & 3\\
2 & 3 & 3 & 1\\
\end{bmatrix}}^{i \ j \ M_{ij} \ p_{ij}}
\qquad \qquad
L^{\text{seq}} = \overbrace{\begin{bmatrix}
1 & 0 & 0 & 4\\
0 & 0 & 0 & 4\\
2 & 0 & 2 & 6\\
1 & 1 & 3 & 5\\
0 & 1 & 2 & 2\\
2 & 1 & 0 & 4\\
0 & 2 & 1 & 6\\
2 & 2 & 1 & 3\\
1 & 2 & 2 & 7\\
2 & 3 & 3 & 1\\
0 & 3 & 3 & 2\\
1 & 3 & 1 & 8\\
\end{bmatrix}}^{i \ j \ M_{ij} \ p_{ij}}
$$
\newline

\noindent The output $L^{\text{seq}}$ can be effectively interpreted as a dispatching list, which can be directly mapped to a schedule as follows:

\begin{enumerate}
    \item Considering $\mathbf{o}'_p=\left[i \ \ l \ M_{il} \ p_{il}\right]$ ($p$-th row of $L^{\text{seq}}$), schedule operation $O_{il}$ to the earliest time such that machine $M_{il}$ is available and, if $l > 0$ (i.e. $O_{il}$ isn't the first operation of $i$-th job) the previous operation $O_{il-1}$ has been executed.
    \item Repeat for all the rows of $L^{\text{seq}}$.
\end{enumerate}

\noindent Mapping $L^{\text{seq}}$ to a JSP solution results in the schedule in \autoref{fig:sch2}.

\subsection{Model architecture}
\label{sec:model_architecture}

\begin{figure}[t]
    \centering
    \includegraphics[width=340px]{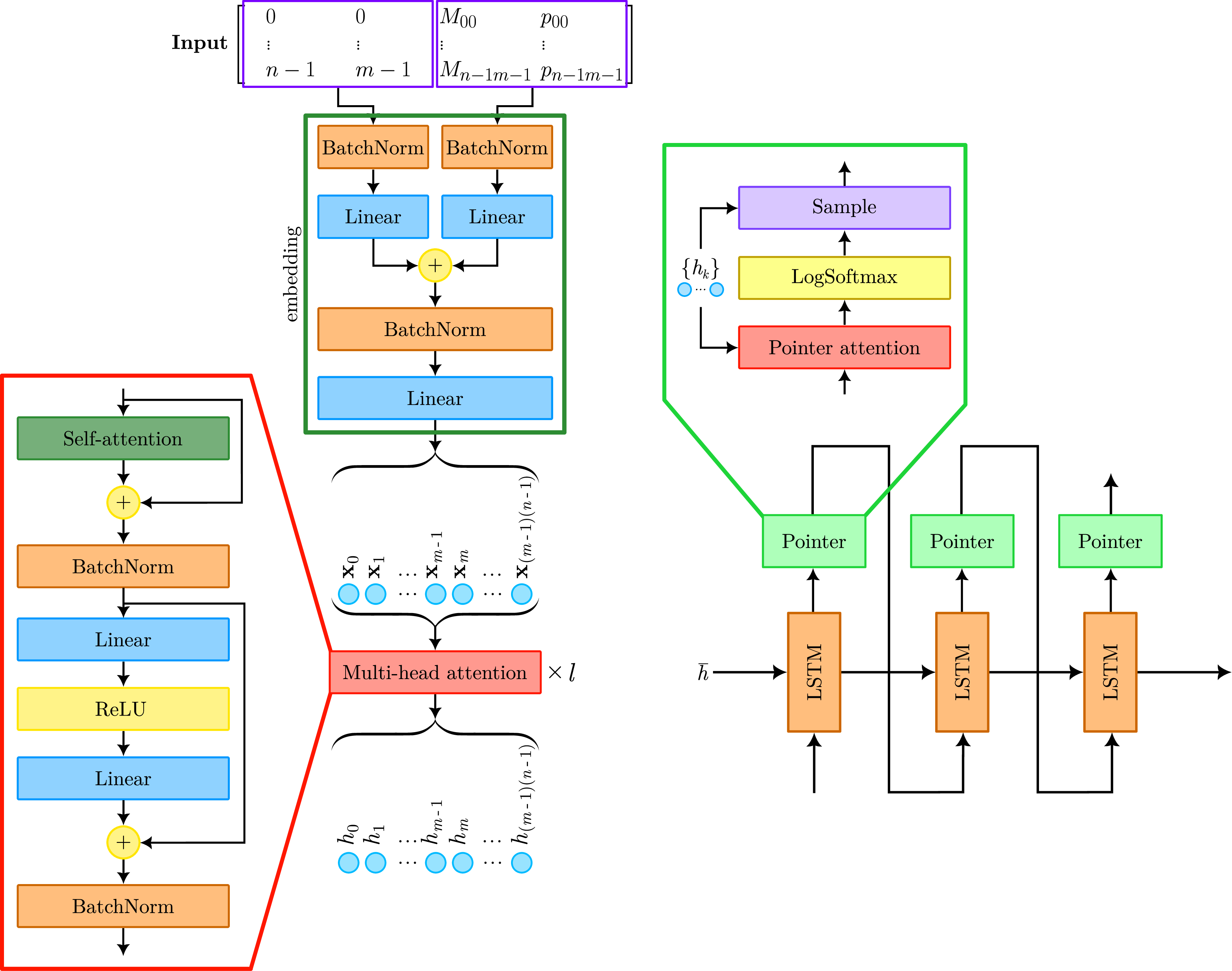}
    \caption{\textit{Left:} Encoder of our model. \textit{Right:} Decoder with pointer mechanism.}
    \label{fig:encoder_decoder_detail}
\end{figure}

Our model is composed of a self-attention-based encoder and a Pointer Network used as decoder (shown in \autoref{fig:encoder_decoder_detail}).

\subsubsection{Encoder} Represented in (\autoref{fig:encoder_decoder_detail}).
The encoder's input is a 3-dimensional tensor $U \in \mathbb{R}^{N \times (nm) \times 4}$ that represents a batch of sequence-encoded instances.
As defined in \autoref{sec:JSP}, $n$ and $m$ are respectively the number of jobs and machines, and $N$ indicates the batch size. 

The first portion of the encoder computes two separate embeddings of each input row, respectively for features $(i, j)$ and $(M_{ij},p_{ij})$, by batch-normalizing and projecting to the embedding dimension $d_h$. After that, the sum of the two vectors is batch-normalized and passed through a linear layer resulting in $X \in \mathbb{R}^{N \times (nm) \times d_h}$.
$X$ is then fed into $l$ multi-head attention layers (we consider $l = 3$).

The output of the encoder is a tensor $H \in \mathbb{R}^{N \times (nm) \times d_h}$ of embeddings $h_k \in \mathbb{R}^{d_h}$, later used as input in the decoder. $\overline{h}$, the average of the these embeddings, is used to initialize the decoder.

\subsubsection{Decoder} Represented in (\autoref{fig:encoder_decoder_detail}),
the decoder is a Pointer Network which generates the policy $\pi_\theta$, a distribution of probability over the rows of the input $S^{\text{seq}}$, via the attention mechanism; during training, the next selected row $\mathbf{o}'_t$ is sampled from it. During evaluation instead, the row with highest probability is selected in a greedy fashion. 
$\pi_\theta$ is defined as follows:

\begin{equation}
\pi_\theta(\mathbf{o}'_t|\mathbf{o}'_0,...,\mathbf{o}'_{t-1}, S^{\text{seq}}) = \text{softmax}\left(\text{mask}(u^t|\mathbf{o}'_0,...,\mathbf{o}'_{t-1})\right)
\end{equation}

\noindent where $u^t$ is the score computed by the Pointer Network's attention mechanism over $S^{\text{seq}}$ input rows. $\text{mask}(u^t|\mathbf{o}'_0,...,\mathbf{o}'_{t-1})$ is a masking mechanism which depends on the sequence partially generated and enforces the constraint in \autoref{def:validJSP}.
\subsubsection{Masking} In order to implement the masking mechanism we use two boolean matrices $M^{\text{sched}}$ and $M^{\text{mask}}$ defined as follows:

\begin{definition}[Boolean matrix  $M^{\text{sched}}$]
Given the $k$-th instance in the batch and the $j$-th operation of the $i$-th job, the element $M^{\text{sched}}_{kp}$ (which refers to $\mathbf{o}_p$, with $p=m \cdot i + j$) is $true$ iff the $j$-th operation has already been scheduled.
\end{definition}

\begin{definition}[Boolean matrix  $M^{\text{mask}}$]
Given the $k$-th instance in the batch and the index $l$ of an operation of the $i$-th job, the element $M^{\text{mask}}_{kp}$ (which refers to $\mathbf{o}_p$, with $p=m \cdot i + l$) is $true$ iff $l > j$, where $j$ is the index of the next operation of the $i$-th job (i.e. scheduling the $l$-th operation would violate \autoref{def:validJSP}).
\end{definition}

\noindent Given $k$-th instance in the batch and $\mathbf{o}_p$ feature vector of the operation scheduled at current time-step, we update $M^\text{sched}$ and $M^\text{mask}$ as follows.

$$
M^\text{sched}_{kp} \gets true, \qquad M^\text{mask}_{kp+1} \gets false
$$

\noindent At current step $t$, the resulting masking procedure of the score associated to input row index $p \in \{0,1,...,(m-1)(n-1)\}$ is the following:

\begin{equation}
\text{mask}(u_p^t|\mathbf{o}'_0,...,\mathbf{o}'_{t-1}) = \left\{
\begin{array}{rl}
    -\infty, & \quad \text{if} \ M^{\text{sched}}_{kp} \ \text{OR} \ M^{\text{mask}}_{kp} \\
    u^t_p, & \quad  \text{otherwise}
\end{array}
\right.
\end{equation}

\begin{figure}
    \centering
    \includegraphics[width=360px]{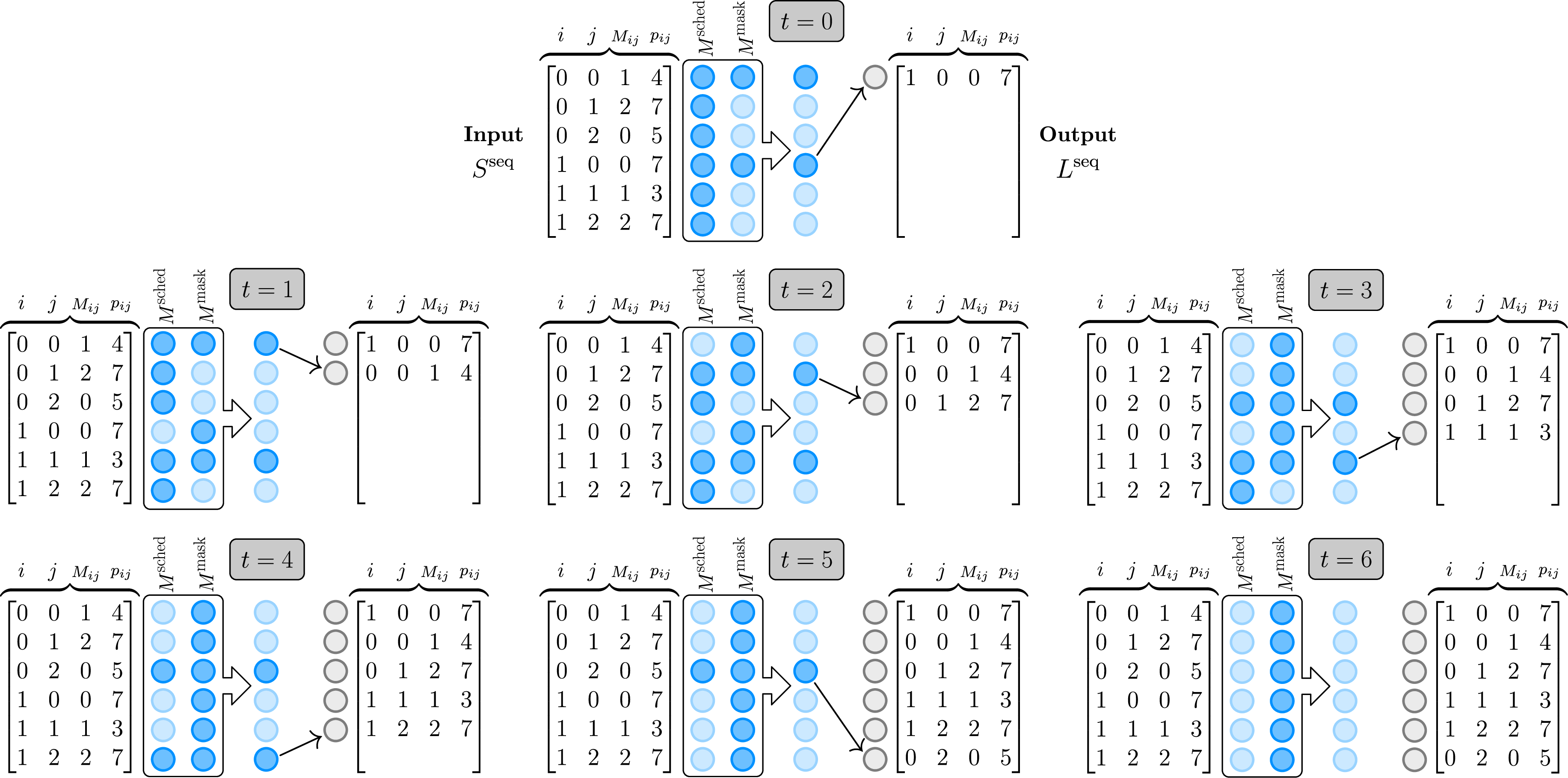}
    \caption{Sequence generation with masking mechanism for the JSP. Light blue circles indicate masked rows and the arrows represent the agent's choices.}
    \label{fig:masking}
\end{figure}

\noindent Masked scores result in a probability close to zero for operations that are already scheduled or cannot be scheduled.
\autoref{fig:masking} shows a possible generation procedure with the masking mechanism just described in order to solve the JSP instance represented in \autoref{tab:jsp_instance_masking}.

\begin{table}[H]
    \centering
    \begin{tabular}{|c|ccc|}
        \hline
        $M_{ij},p_{ij}$ & $O_{*0}$ & $O_{*1}$ & $O_{*2}$ \\
        \hline
         $J_0$ & (1, 4) & (2, 7) & (0, 5) \\
        $J_1$ & (0, 7) & (1, 3) & (2, 7) \\
         \hline
    \end{tabular}
    \caption{Example of a $2 \times 3$ JSP instance.}
    \label{tab:jsp_instance_masking}
\end{table}

\subsubsection{Training algorithm} The network is trained with REINFORCE \cite{williams1992simple} described in \autoref{sec:policy_gradient} using the Adam optimizer. We use the following form of the policy gradient:

\begin{equation}
\nabla_\theta L(\pi_\theta) = \mathop{\mathbb{E}} \left[ \left( C_{max}(L^{\text{seq}}) - b(S^{\text{seq}}) \right)\nabla_\theta \log P_\theta(L^{\text{seq}}|S^{\text{seq}}) \right]
\end{equation}

\noindent where $P_\theta(L^\text{seq}|S^{\text{seq}})=\prod_{t=0}^{nm-1}\pi_\theta(\mathbf{o}'_t|\mathbf{o}'_0,...,\mathbf{o}'_{t-1},S^\text{seq})$ is the probability of the solution $L^\text{seq}$ and $b(S^{\text{seq}})$ is the greedy rollout baseline. 
After each epoch, the algorithm updates the baseline with the optimized policy's weights if the latter is statistically better. This is determined by evaluating both policies on a 10000 samples dataset and running a paired t-test with $\alpha=0.05$ (see \cite{kool2019attention} for the detailed explanation). The periodic update ensures that the policy is always challenged by the best model, hence the reinforcement of actions is effective.  From a RL perspective, $-C_{max}(L^{\text{seq}})$ is the reward of the solution --- lower makespan implies higher reward.
After training, the active search approach \cite{bello2017neural} is applied.


\subsubsection{Solving related scheduling problems} Our method represents a general approach to scheduling problems and, once trained on JSP instances, it can also solve the Flow Shop Problem. The Open Shop Problem can also be solved

\vspace{20px}
\textit{...OMISSIS...}

\subsection{Experiments and results}

In this section we present our experiments and results. We consider four JSP settings: $6\times 6$, $10\times 10$, $15\times 15$ and $30\times 20$. After hyperparameter tuning, we set the learning rate to $10^{-5}$ and gradient clipping to $0.5$ in order to stabilize training. At each epoch, the model processes a dataset generated with the well-known Taillard's method \cite{taillard1993benchmarks}. \autoref{tab:training_config} sums up training configurations for every experiment.

\begin{figure}[h]
    \centering
    \includegraphics[width=360px]{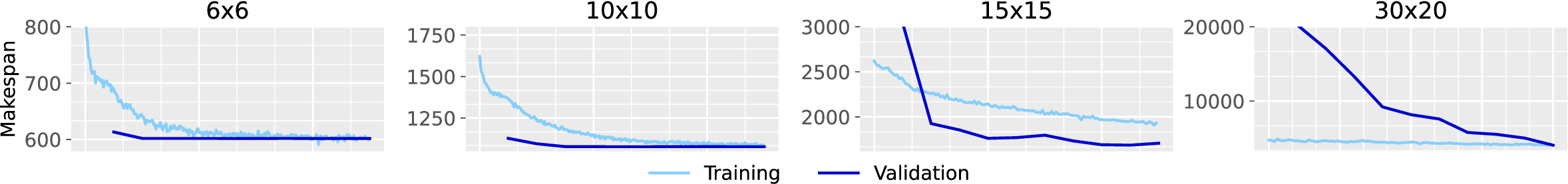}
    \caption{Training and Validation curves for different JSPs.}
    \label{fig:plots}
\end{figure}

During training we note the average cost every 50 batches and the validation performance at the end of every epoch. Validation rollouts are done in a greedy fashion, i.e. by choosing actions with maximum likelihood. Training and validation curves are represented in \autoref{fig:plots}.

\begin{table}
    \makegapedcells
    \centering
    \setcellgapes{1pt}%
    \begin{tabular}{>{\centering}p{1.5cm}|>{\centering}p{1.8cm}>{\centering}p{1.5cm}ccc}
        \specialrule{.2em}{.1em}{.1em} 
        Size & Epoch size & N° epochs & Batch size & GPU(s)$^*$ & Duration \\
        \hline
        $6 \times 6$ & 640000 & 10 & 512 & Titan RTX & 30m \\
        $10 \times 10$ & 640000 & 10 & 512 & RTX A6000 & 1h 30m  \\
        $15 \times 15$ & 160000 & 10 & 256 & RTX A6000 & 1h 30m \\
        $30 \times 20$ & 16000 & 10 & 32 & Titan RTX & 1h 45m  \\
        \hline
    \end{tabular}
    \caption{Training configurations for all the experiments. $^*$ Nvidia GPUs have been used.}
    \label{tab:training_config}
\end{table}

\vspace*{-30px}

\vspace*{-20px}

\subsubsection{Comparison with concurrent work}

As already said in the \hyperref[sec:intro]{Introduction}, we compare our results with the work from Zhang et al. \cite{zhang2020learning}, and with a set of largely used dispatching rules: \textit{Shortest Processing Time} (SPT), \textit{Most Work Remaining} (MWKR), \textit{Most Operations Remaining} (MOPNR), \textit{minimum ratio of Flow Due Date to most work remaining} (FDD). 

\autoref{tab:jsp_results} shows the testing results obtained applying our technique on 100 instances generated by Zhang et al. with the Taillard's method.

We compare each solution with the optimal one obtained with Google OR-Tools' \cite{perron2011operations} solver; in the last column we report the percentage of instances for which OR-Tools returns optimal solutions in a limited computation time of 3600 seconds. 

The column JSP settings shows the average makespan over the entire test dataset and the gap between $\overline{C}_{max}$ (the average makespan of heuristic solutions) and $\overline{C}^*_{max}$ (the average makespan of the optimal ones), defined as $\overline{C}_{max}/\overline{C}^*_{max}-1$.

\vspace*{-5px}

\begin{table}[H]
    \makegapedcells
    \centering
    \renewcommand{\arraystretch}{1.5}
    \setcellgapes{4pt}%
    \resizebox{\textwidth}{!}{
    \begin{tabular}{cc|cccccc|c}
        \specialrule{.2em}{.1em}{.1em} 
        \multicolumn{2}{>{\centering} p{2.7cm}|}{JSP settings} & SPT & MWKR & FDD & MOPNR & Zhang \cite{zhang2020learning} & Ours & Opt. Rate(\%) \\
        \hline
        $6 \times 6$ & \makecell{$\overline{C}_{max}$ \\ Gap } & \makecell{691.95 \\ 42.0\% } & \makecell{656.95 \\ 34.6\% } & \makecell{604.64 \\ 24.0\% } & \makecell{630.19 \\ 29.2\% } & 
        \makecell{574.09 \\ 17.7\% } & 
        \makecell{\textbf{495.92} \\ \textbf{1.7\%} } & 100\% \\
        \hline
        $10 \times 10$ & \makecell{$\overline{C}_{max}$ \\ Gap } & \makecell{1210.98 \\ 50.0\% } & \makecell{1151.41 \\ 42.6\% } & \makecell{1102.95 \\ 36.6\% } & \makecell{1101.08 \\ 36.5\% } &
        \makecell{988.58 \\ 22.3\%} & 
        \makecell{ \textbf{945.27} \\ \textbf{16.9\%} } & 100\% \\
         \hline
        $15 \times 15$ & \makecell{$\overline{C}_{max}$ \\ Gap } & \makecell{1890.91 \\ 59.2\%} & \makecell{1812.13 \\ 52.6\% } & \makecell{1722.73 \\ 45.1\% } & \makecell{1693.33 \\ 42.6\% } & 
        \makecell{\textbf{1504.79} \\ \textbf{26.7\%}} & 
        \makecell{1535.14 \\ 29.3\% } & 99\% \\
         \hline
        $30 \times 20$ & \makecell{$\overline{C}_{max}$ \\ Gap } & \makecell{3208.69 \\ 65.3\% } & \makecell{3080.11 \\ 58.7\% } & \makecell{2883.88 \\ 48.6\%} & \makecell{2809.62 \\ 44.7\%} & 
        \makecell{\textbf{2508.27} \\ \textbf{29.2\%}} & 
        \makecell{2683.05 \\ 38.2\%} & 12\% \\
        \hline
    \end{tabular}
    }
    \caption{Results over different JSP settings.}
    \label{tab:jsp_results}
\end{table}

\vspace*{-20px}

\noindent From \autoref{tab:jsp_results} we can see that our model greatly outperforms the traditional dispatching rules even by a margin of $71\%$ with respect to SPT. When compared to \cite{zhang2020learning} our model is superior in performance in the $6 \times 6$ and $10 \times 10$ cases, while having similar results in the $15 \times 15$ JSPs, and sligthly underperforming in the $30 \times 20$.
Speculating about the drop in performance of our solution in the biggest settings (i.e. $30 \times 20$ JSPs) we think it could be due to the following reasons:

\begin{itemize}
    \item Larger JSP instances are encoded by longer sequences: like traditional RNNs and transformers, our model tends to have a suboptimal representation of the input if the sequence is exceedingly long.
    \item As mentioned before, for execution time reasons we reduce the number of instances and examples in each batch: this implies a gradient estimate with higher variance, hence a potentially unstable and longer learning.
\end{itemize}

\subsubsection{Improving Active Search through Efficient Active Search}
\noindent Efficient Active Search (EAS) is a technique introduced in a recent work by Hottung et al. \cite{hottung2022efficient} that extends and substantially improves active search, achieving state-of-the-art performance on the TSP, CVRP and JSP. The authors proposed three different techniques, EAS-Emb, EAS-Lay and EAS-Tab, all based on the idea of performing active search while adjusting only a small subset of model parameters. EAS-Emb achieves the best performance and works by keeping all model parameters frozen while optimizing the embeddings. As pointed out in \cite{hottung2022efficient}, this technique can be applied in parallel to a batch of instances, greatly reducing the computing time. Here we present a preliminary attempt to extend our method applying EAS-Emb and we test it on the 10x10 JSP. \autoref{tab:eas_results} shows that our model greatly benefits from the use of EAS-Emb, although underperforming Hottung et al.'s approach.

\vspace*{-10px}

\begin{table}[H]
    \makegapedcells
    \centering
    \setcellgapes{1pt}%
    \begin{tabular}{cc|>{\centering\arraybackslash}p{3cm}>{\centering\arraybackslash}p{3cm}}
        \specialrule{.2em}{.1em}{.1em} 
        \multicolumn{2}{>{\centering} p{2.7cm}|}{JSP settings} & Hottung et al. \cite{hottung2022efficient} & Ours+EAS-Emb \\
        \hline
        $10 \times 10$ & \makecell{$\overline{C}_{max}$ \\ Gap } & \makecell{837.0 \\ 3.7\%} & \makecell{864.9 \\ 7.2\%} \\
         \hline
    \end{tabular}
    \caption{Efficient Active Search: comparison results}
    \label{tab:eas_results}
\end{table}

\vspace*{-40px}

\section{Conclusions}
In this work we designed a Sequence to Sequence model to tackle the JSP, a famous Combinatorial Optimization problem, and we demonstrated that it is possible to train such architecture with a simple yet effective RL algorithm. Our system automatically learns dispatching rules and relies on a specific masking mechanism in order to generate valid schedulings. Furthermore, it is easy to generalize this mechanism for the Flow Shop Problem and the Open Shop Problem with none or slight modifications.
Our solution beats all the main traditional dispatching rules by great margins and achieve better or state of the art performance on small JSP instances. 

For future works we plan to improve the performance of our method on larger JSP instances exploiting EAS-based approaches. Besides, although this work is mostly concerned with evaluating a Deep RL-based paradigm for combinatorial optimization, the idea of hybridizing these techniques with more classical heuristics remain viable. At last, one promising idea would be to improve our method using Graph Neural Networks as encoders. Graph-based models could produce more refined embeddings exploiting the disjunctive graph representation of scheduling problem instances.

\subsubsection{Acknowledgements} The activity has been partially carried on in the context of the Visiting Professor Program of the Gruppo Nazionale per il Calcolo Scientifico (GNCS) of the Italian Istituto Nazionale di Alta Matematica
(INdAM).

%
%
%
\bibliographystyle{splncs04}

\bibliography{mybibliography}

\end{document}